\documentclass{article}

 \usepackage[preprint]{neurips_2026}

\usepackage[utf8]{inputenc} 
\usepackage[T1]{fontenc}    
\usepackage{hyperref}       
\usepackage{url}            
\usepackage{booktabs}       
\usepackage{amsfonts}       
\usepackage{nicefrac}       
\usepackage{microtype}      
\usepackage[table]{xcolor}         

\usepackage[pdftex]{graphicx}

\title{ReorgGS: Equivalent Distribution Reorganization for 3D Gaussian Splatting}

\author{
    Luchao Wang \; \; 
    Kaimin Liao \; \; 
    Qian Ren \; \; 
    Hua Wang \\
    \textbf{
    Zhi Chen \; 
    Yaohua Tang \; } 
     \\
    \texttt{tangyaohua28@gmail.com}
}

\begin{document}

\maketitle

\begin{abstract}

A converged 3D Gaussian Splatting (3DGS) model may approximate the target scene while remaining poorly parameterized for further optimization. We identify this failure mode as \emph{parameterization degeneration}: high-opacity floaters attenuate gradients to true surfaces through alpha compositing, and redundant overlapping clusters create strongly coupled parameter blocks with nearly collinear Jacobian responses. These effects explain why continued optimization can plateau even when the model still contains removable artifacts. We propose ReorgGS, an equivalent distribution reorganization method for converged 3DGS models. ReorgGS treats the existing Gaussian set as an empirical probability field, resamples centers from it, estimates local anisotropic covariances with kNN, initializes low opacity, and continues optimization with the original 3DGS renderer and loss. Unlike opacity reset, which only rescales opacity on the old overlap graph, ReorgGS rebuilds centers, covariances, and visibility structure, thereby changing the graph itself. Our analysis shows that distributional equivalence is not optimization equivalence. The reorganized model preserves scene support while improving gradient accessibility under alpha compositing and reducing opacity-weighted overlap, thereby weakening  local parameter coupling during subsequent optimization. Under the same additional optimization budget, ReorgGS improves fitting quality at a fixed Gaussian count, suppresses persistent floaters, and reduces rendering overhead from redundant overlap.


\begin{figure}[t]
  \centering 
    \includegraphics[width=1.0\linewidth]{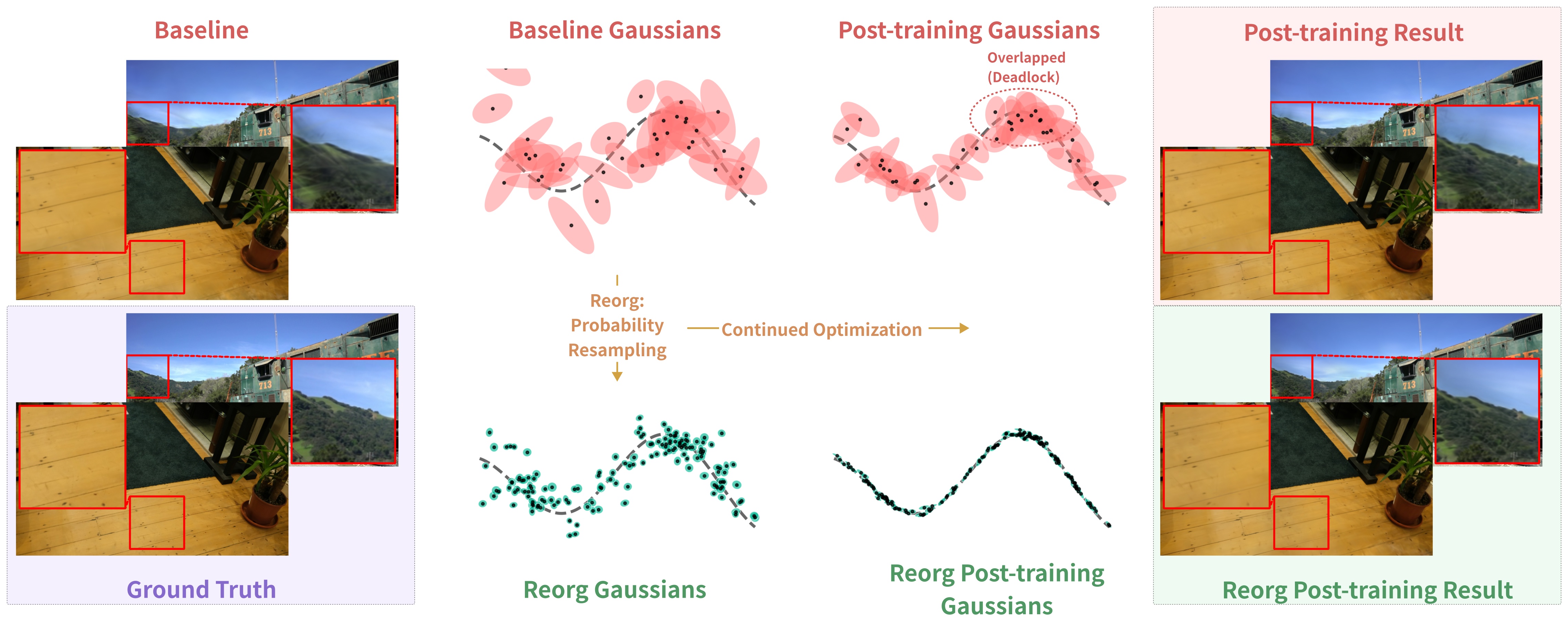}
  \caption{Reorganization unlocks the full potential of post-training. Given a suboptimal baseline (e.g., LiteGS, top-left), directly applying post-training methods (top row) fails to resolve the underlying thick and messy geometric topology. This overlapped configuration causes gradient deadlock during optimization, leaving fine details persistently blurred in the final rendering (top-right). In contrast, our ReorgGS (bottom row) first resamples the baseline's chaotic probability field into discrete, unoccluded primitives. This fundamental topological reset smooths the optimization landscape, allowing the same post-training process to explicitly align the points to the local manifold and easily recover high-fidelity details (bottom-right) that closely match the Ground Truth (bottom-left).}
  \label{fig:teaser}
\end{figure}

\end{abstract}

\section{Introduction}

3D Gaussian Splatting (3DGS)~\citep{kerbl20233d} represents scenes with anisotropic 3D Gaussians and renders them via differentiable alpha compositing, achieving strong reconstruction quality, fast training, and efficient rendering. The standard 3DGS pipeline builds this representation through local, history-dependent heuristic operations—such as gradient-based densification, cloning, splitting, and pruning, alongside periodic opacity resets. While these operations rapidly expand scene coverage, they also inevitably rigidify the occlusion order, overlap relations, and parameter-coupling structure of the Gaussian set.

This trajectory dependence induces a late-stage optimization bottleneck—one rooted not in representation capacity, but in parameter topology. In many converged models, training views are rendered reasonably well, yet high-opacity floaters persist in free space, redundant Gaussians repeatedly explain the same local residuals, and directly continuing optimization yields negligible gains. The root cause lies in the non-uniqueness of the 3DGS representation: different Gaussian sets can project to nearly identical images, yet possess drastically different visibility structures and optimization landscapes. Through alpha compositing, a high-opacity floater can act as an opaque barrier, attenuating gradients to true surfaces behind it. Likewise, a redundant overlapping cluster forces multiple parameter blocks to respond to the same pixel residual, causing severe collinearity. We term this state \emph{degenerate parameterization}: the model is distributionally adequate, but optimization-degenerate. This perspective elegantly unifies three traditionally disparate symptoms: floaters correspond to ray-level gradient truncation, redundant overlap induces strong local coupling, and optimization plateauing is the inevitable consequence of these structural barriers. This leads to the central question of our work:

\begin{quote}
If a converged 3DGS model already approximately represents a scene, can we replace it—without increasing the Gaussian budget—with a distributionally equivalent but strictly easier-to-optimize parameterization?
\end{quote}

We propose ReorgGS to answer this question. Given a converged Gaussian set \(\mathcal{G}^p\), ReorgGS interprets it as an empirical Gaussian-mixture probability field. We resample centers from this field, estimate local anisotropic covariances using k-nearest neighbors, and initialize a new, low-opacity Gaussian set \(\mathcal{G}^q\). This approach fundamentally distinguishes ReorgGS from standard opacity resets. A standard reset merely applies \(\alpha_i \leftarrow \alpha_0\) while preserving the original centers, covariances, and old overlap graph; it mitigates weights on a flawed topology, but cannot escape the topology itself. ReorgGS, instead, applies low opacity to newly sampled centers with geometrically re-estimated shapes, thereby explicitly dismantling and rebuilding the visibility structure that dictates occlusion and parameter coupling. Our contributions are summarized as follows:

\begin{enumerate}
\item \textbf{We identify parameterization degeneration as a critical late-stage bottleneck in 3DGS.} We demonstrate that persistent floaters, redundant overlap, and optimization plateaus are symptoms of an unfavorable optimization coordinate system, rather than simply a lack of capacity or training time.
\item \textbf{We introduce ReorgGS, an equivalent distribution reorganization method.} ReorgGS preserves the learned scene support while resampling centers, re-estimating anisotropic covariances, and resetting opacity. This fundamentally improves visibility structure and reduces local coupling without altering the renderer, loss function, or Gaussian budget.
\item \textbf{We analyze and validate the optimization dynamics of reorganization.} We mathematically and empirically show that distributional equivalence does not imply optimization equivalence. ReorgGS improves gradient accessibility and reduces parameter coupling, delivering up to 0.5 dB higher training PSNR and 0.14 dB higher test PSNR over direct continuation, along with a 12.8\% rendering speedup in representative settings.
\end{enumerate}

\section{Related Work}

\textbf{Representational Variants of 3DGS.} 3DGS parameterizes scenes with depth-sorted anisotropic primitives~\citep{kerbl20233d}. To improve geometric fidelity, numerous variants alter the fundamental representation via neural anchors~\citep{lu2024scaffold}, planar constraints~\citep{huang20242d}, geometry extraction~\citep{yu2024gaussian}, or anti-aliasing filters~\citep{yu2024mip}. While these architectural shifts confirm the importance of primitive organization, they achieve it by modifying the underlying mathematical formulation. ReorgGS takes an orthogonal path: we demonstrate the standard 3DGS primitive is strictly sufficient. Instead of altering the representation, we reorganize a degenerate Gaussian set into a well-conditioned parameterization to unlock superior optimization without changing the renderer.

\textbf{Heuristic Density Control.} Standard 3DGS relies on heuristic densification and opacity resets to shape the Gaussian distribution~\citep{kerbl20233d}. Subsequent works refine these micro-level operations via advanced gradient signals, steepest-descent, localized point management, or dedicated floater-free frameworks~\citep{cheng2024gaussianpro,zhang2024pixel,ye2024absgs,yang2024gaussian,wang2025steepest,chen2025dashgaussian,wang2025stablegs}. However, these methods primarily govern \emph{incremental growth} and remain greedy. They struggle to undo structural damage (e.g., redundant overlap, deadlocks) accumulated over long training trajectories. Rather than patching local growth heuristics, ReorgGS executes a global topological reset to dismantle unfavorable occlusion orders and parameter collinearity in the post-plateau regime.

\textbf{Compact 3DGS and Probabilistic Sampling.} Extensive research reduces 3DGS overhead via pruning, quantization, distillation, or entropy coding~\citep{niedermayr2024compressed,fan2024lightgaussian,girish2024eagles,lee2024compact,fang2024mini,mallick2024taming,wang2024contextgs,dai2025efficient,lee2025compression}. These methods universally treat a converged Gaussian set as a static object to be compressed. In contrast, ReorgGS treats it as a \emph{flawed optimization coordinate system}, actively improving gradient accessibility to break fitting ceilings under restricted point budgets. Closest to our probabilistic perspective is 3DGS-MCMC~\citep{kheradmand20243d}, which injects stochastic exploration into the training loop. While MCMC performs continuous micro-exploration \emph{during} optimization, ReorgGS intervenes at the macro level: we halt optimization, resample the degenerate probability field into a clean, unoccluded state, and provide a drastically flattened loss landscape for the optimizer to resume.

\begin{figure}[t]
  \centering 
    \includegraphics[width=1.0\linewidth]{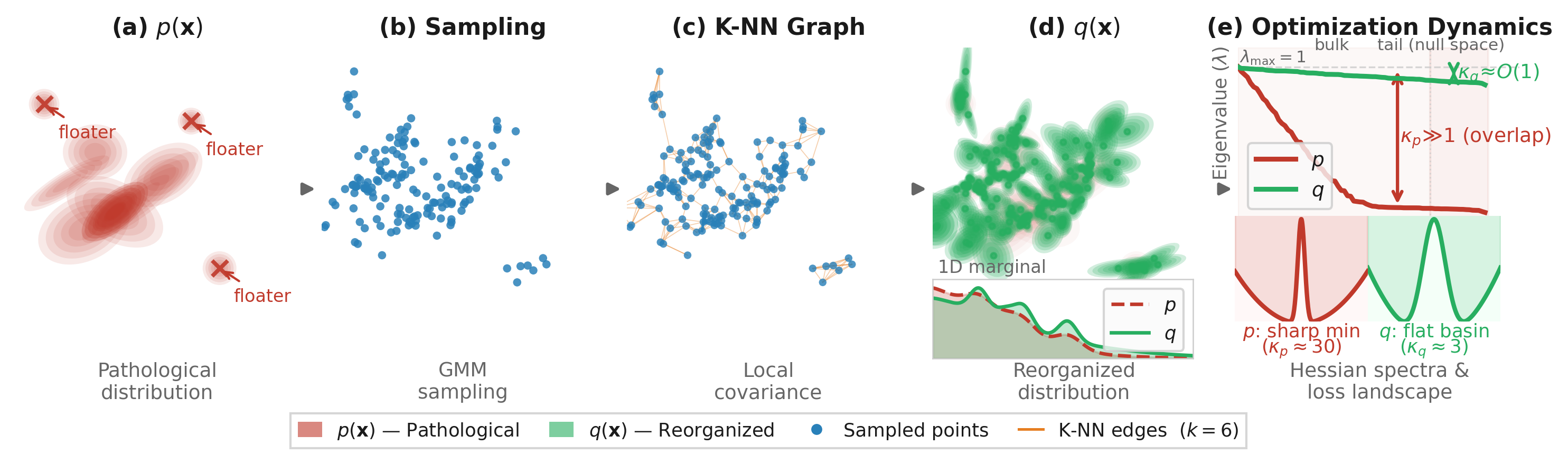}
  \caption{Method pipeline and optimization dynamics of ReorgGS. (a) Vanilla 3DGS often degenerates into a pathological distribution $p(x)$ with severe spatial overlap. (b) Our $\alpha$-driven equivalent resampling discards the spatial volume term, naturally filtering out large-scale floating artifacts. (c, d) Via kNN local manifold alignment, we reconstruct the covariance to form a healthy continuous field $q(x)$ with bounded overlap. (e) Schematic illustration of Hessian Spectrum Analysis: In the vanilla distribution $p$, severe physical overlap causes heavy occlusion, forcing a massive amount of eigenvalues to collapse from the bulk into the tail (null space), leading to a catastrophic condition number ($\kappa_p \gg 1$) and sharp deadlocks. Conversely, ReorgGS bounded overlap effectively eliminates the null space, keeping the tail eigenvalues stable ($\kappa_q \approx O(1)$). This preconditions the loss landscape, smoothly transforming sharp deadlocks into optimizable flat minima.}
  \label{fig:pipeline}
\end{figure}

\section{Method}
\subsection{From 3DGS Parameters to a Reorganized Distribution}

A 3DGS model represents a scene by a set of Gaussians \(\mathcal{G}=\{g_i\}_{i=1}^{N}\). Each primitive has a center \(\mu_i\), covariance \(\Sigma_i\), opacity \(\alpha_i\), and color parameters \(C_i\), with spatial response $G_i(x)=\exp\left(-\frac{1}{2}(x-\mu_i)^T\Sigma_i^{-1}(x-\mu_i)\right)$. ReorgGS starts from a model that has converged or reached a plateau. Rather than treating its primitives as fixed optimization variables, we interpret their spatial support as an unnormalized empirical Gaussian-mixture field: 
$P(x) \approx \sum_{i=1}^{N}\alpha_i\mathcal{N}(x\mid \mu_i,\Sigma_i).$

This field is not introduced to change the rendering model. It is a compact way to separate what the current model has learned from how it is parameterized. Many Gaussian sets can induce nearly the same spatial support while having very different opacity patterns, overlap relations, and visibility orders. Standard densification, pruning, and opacity reset gradually construct one such parameterization, but the final arrangement need not be a good coordinate system for further optimization. The importance of this distinction comes from alpha compositing. For a ray or pixel \(r\), 3DGS renders $C(r)=\sum_{i=1}^{N}T_i(r)a_i(r)C_i(r), T_i(r)=\prod_{j<i}(1-a_j(r))$, where \(a_i(r)\) is the effective projected opacity and \(T_i(r)\) is the accumulated transmittance before the \(i\)-th Gaussian. A high-opacity floater or a redundant foreground cluster can make \(T_i(r)\) small, starving the true surface behind it of gradient. Thus, continued optimization may fail not because the scene support is absent, but because the existing support is organized with an unfavorable visibility and overlap structure.

Direct continuation keeps this structure and updates the old variables in place, $\mathcal{G}_i \leftarrow \mathcal{G}_i-\eta\nabla_{\mathcal{G}_i}\mathcal{L}.$ ReorgGS instead constructs a new Gaussian set \(\mathcal{G}'=\{g'_j\}_{j=1}^{M}\), with parameters \(\{\mu'_j,\Sigma'_j,\alpha'_j,C'_j\}_{j=1}^{M}\), whose induced field $q(x)=\sum_{j=1}^{M}\alpha'_j\mathcal{N}(x\mid \mu'_j,\Sigma'_j)$ approximately preserves the support of \(P(x)\) while changing the centers, covariances, opacity initialization, and overlap graph. The goal is therefore not only distributional similarity, \(q\approx P\), but optimization improvement: the reorganized set should expose gradients more evenly and reduce local parameter coupling during the subsequent 3DGS optimization.

\subsection{Algorithm}

Given a 3DGS model \(\{\mu_i,\Sigma_i,\alpha_i,C_i\}_{i=1}^{N}\) that has reached the middle-to-late stage of training, converged, or entered a plateau, ReorgGS constructs a new Gaussian set as follows.

\paragraph{Step 1. GMM modeling.}
We view the current Gaussian set as an empirical Gaussian-mixture field and define its normalized form as $\hat{p}(x)=\frac{P(x)}{Z}$, where $Z=\int P(x)dx$. In implementation, the continuous integral \(Z\) does not need to be computed explicitly. Sampling only requires the discrete mixture weights of the components.

\paragraph{Step 2. Categorical sampling.}
We first sample a latent variable $z_j\in\{1,\ldots,N\}$ from a categorical distribution $P(z_j=k)=\frac{\alpha_k}{\sum_{l=1}^{N}\alpha_l}$.  Thus, Gaussian components with higher opacity are more likely to be selected. We use only \(\alpha_k\) as the categorical weight and do not multiply it by the Gaussian volume. The reason is that, in degenerate 3DGS optimization, abnormally large Gaussians often correspond to blurred regions or artifacts. Volume-weighted sampling could further amplify such large-scale abnormal structures. We therefore ignore volume in the categorical probability to reduce this sampling bias.

\paragraph{Step 3. Gaussian coordinate sampling.}
After choosing category \(z_j=k\), we sample a new center from the corresponding 3D Gaussian $x_j\sim\mathcal{N}(\mu_k,\Sigma_k).$ In practice, we use the reparameterized form $x_j=\mu_k+L_k\epsilon$, where $\epsilon\sim\mathcal{N}(0,I)$ and \(L_k\) is the Cholesky factor of \(\Sigma_k\). Since 3DGS parameterizes \(\Sigma_k=R_kS_kS_k^TR_k^T\), one can also directly use \(R_kS_k\) as the sampling transform. Through this procedure, ReorgGS regenerates centers from the continuous spatial distribution induced by the old model, rather than simply copying old Gaussian positions.

\paragraph{Step 4. Anisotropic initialization.}
Sampled centers alone are not sufficient for an effective 3DGS initialization. To make the new Gaussians quickly adapt to local geometry, we perform a kNN query for each sampled point \(x_j\) within the new point set \(\{x_j\}_{j=1}^{M}\), obtaining its \(k\) nearest neighbors $
\{x_{j_1},\ldots,x_{j_k}\}.$ We then estimate the new covariance using the local second moment: $\Sigma'_j=\frac{1}{k}\sum_{l=1}^{k}(x_{j_l}-x_j)(x_{j_l}-x_j)^T$. The kNN covariance provides a natural anisotropic initialization.

\paragraph{Step 5. Constructing the reorganized distribution \(q(x)\).}
Using sampled point \(x_j\) as the center and local covariance \(\Sigma'_j\) as the shape, we construct the new Gaussian set. We set $\mu'_j=x_j, \alpha'_j=\alpha_0, $ where \(\alpha_0\) is a small opacity value (default 0.01). The spatial scale of each reorganized Gaussian is naturally determined by \(\Sigma'_j\), without additional volume normalization. Subsequent optimization continues to adjust scale, rotation, and opacity to fit the image reconstruction objective. The appearance parameter \(C'_j\) is inherited from the nearest old Gaussian: $\mathrm{NN}(x_j)=\arg\min_i\|x_j-\mu_i\|_2, C'_j=C_{\mathrm{NN}(x_j)}$.

\paragraph{Step 6. Recovering scale and rotation.}
To write \(\Sigma'_j\) back into the standard 3DGS parameterization, we perform eigendecomposition $\Sigma'_j = U_j \Lambda_j U_j^T$, set $R'_j = U_j$ and $S'_j = \sqrt{\Lambda_j}$, so that $\Sigma'_j = R'_j S'_j (S'_j)^T (R'_j)^T$. After obtaining \((\mu'_j,S'_j,R'_j,\alpha'_j,C'_j)\), we continue training with the original 3DGS renderer, loss function, and optimization strategy. Since the Gaussian set has been rebuilt, the optimizer state is reinitialized for the new parameters.

\subsection{Optimization Properties of the Reorganized Distribution}

ReorgGS is not only a distribution approximation. Preserving the scene support, i.e., making \(q(x)\) close to \(P(x)\), does not imply that the two Gaussian sets have the same visibility, gradients, or local curvature. The benefit of ReorgGS comes from reorganizing four coupled structures: opacity-density allocation, local overlap, alpha-compositing transmittance, and covariance alignment. Appendix~\ref{app: tp} provides the detailed analysis.

\textbf{1. From opacity values to spatial density.}
Let \(p(x)\) be the normalized spatial distribution of the original model, and let \(q(x)\) be the mixture induced by resampled centers and kNN covariances. Under standard kNN density-estimation assumptions, Appendix~\ref{app: deco} gives $\mathbb{E}[q(x)]\to p(x), \mathrm{Var}[q(x)]=O(1/k)$. With constant low opacity \(\alpha_0\) (set to 0.01, see Sec. 3.2), the reorganized occupancy field satisfies $\mathbb{E}[\hat{F}(x)]\approx M\alpha_0p(x).$ Thus, ReorgGS transfers scene occupancy from large per-primitive opacity to spatial sample density: dense regions receive more primitives, sparse regions fewer, while each primitive starts with low opacity. This preserves support while reducing absolute ray occlusion.

\textbf{2. Spatial demixing and bounded overlap.} 
Standard continued optimization keeps the old overlap graph created by densification. ReorgGS instead sets covariance scales from kNN neighborhoods. If the local point density is \(\rho(x)=Mp(x)\), the effective number of local overlaps satisfies (Appendix~\ref{app: bokr}): $K_q(x) \approx \rho(x)\cdot\frac{k}{\rho(x)} = O(k).$ Hence local coverage is controlled by the neighborhood size rather than by the number of redundant clones in the original cluster. 

\textbf{3. Gradient accessibility.}
Under alpha compositing, a primitive behind foreground components receives gradients scaled by $T_i=\prod_{j<i}(1-\alpha_j).$ High-opacity floaters or many redundant foreground primitives can drive \(T_i^p\to0\), starving true surfaces of gradient. ReorgGS initializes low opacity and bounds foreground overlap; if \(m_q=O(k)\), then $T_i^q\ge(1-\alpha_0)^{m_q} \approx 1-m_q\alpha_0 > 0.$ The improvement therefore comes from a new visibility structure, not from merely lowering opacity on the old one (Appendix~\ref{app: gaac}).

\textbf{4. Local manifold alignment and improved conditioning.}
Redundant Gaussians with inconsistent scales and orientations can explain the same residual with nearly collinear Jacobian responses, increasing off-diagonal coupling in the Gauss--Newton Hessian or Fisher Information Matrix. kNN covariance initialization aligns nearby primitives with the local tangent-normal structure. If \(\Sigma_i\approx\Sigma_{\mathrm{manifold}}\), then $H_{\mathrm{local}} \propto \sum_i\Sigma_i^{-1} \approx K\,U\Lambda^{-1}U^T$, where \(U\) is the local eigenbasis. ReorgGS effectively eliminates the null space and transforms sharp deadlocks into optimizable flat minima (Appendix~\ref{app: hflma}).

\begin{figure}[t]
  \centering 
    \includegraphics[width=1.0\linewidth]{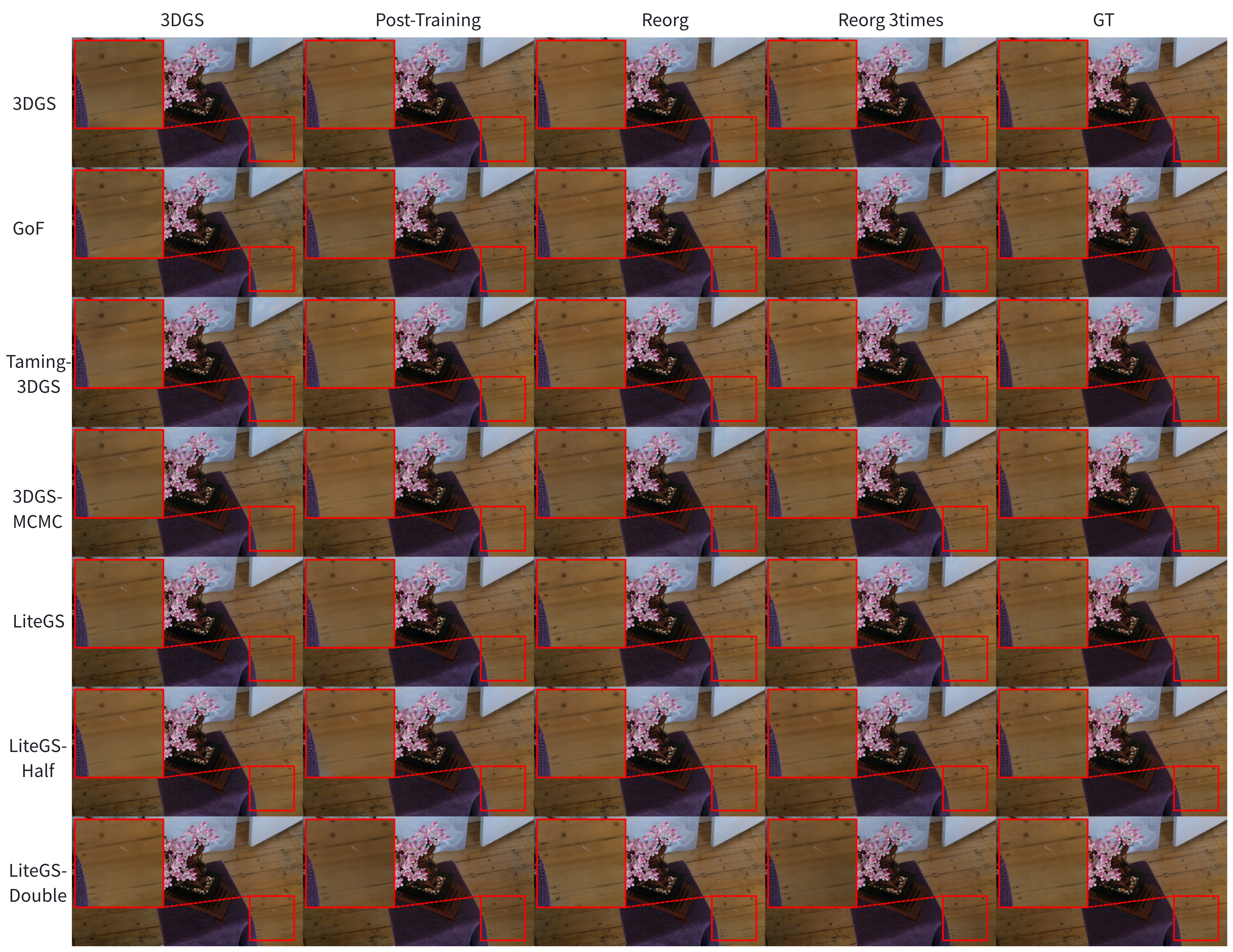}
  \caption{Plug-and-play capability and qualitative comparisons across diverse 3DGS baselines. We evaluate the universality of our Reorg mechanism on various initializations (rows), including standard 3DGS, GoF, Taming-3DGS, and LiteGS variants. Columns from left to right: Original baseline, standard Post-Training, our Reorg (single pass), Reorg applied iteratively (Reorg $\times$3), and Ground Truth (GT). As highlighted in the red insets (wood grain textures), standard post-training struggles to resolve the blurry artifacts caused by suboptimal topology. In contrast, our Reorg consistently unlocks sharp, high-fidelity fine details across all tested baselines. Furthermore, applying Reorg iteratively further refines the geometry, achieving visual quality highly consistent with the GT.}
  \label{fig:result}
\end{figure}

\begin{figure}[t]
  \centering 
    \includegraphics[width=1.0\linewidth]{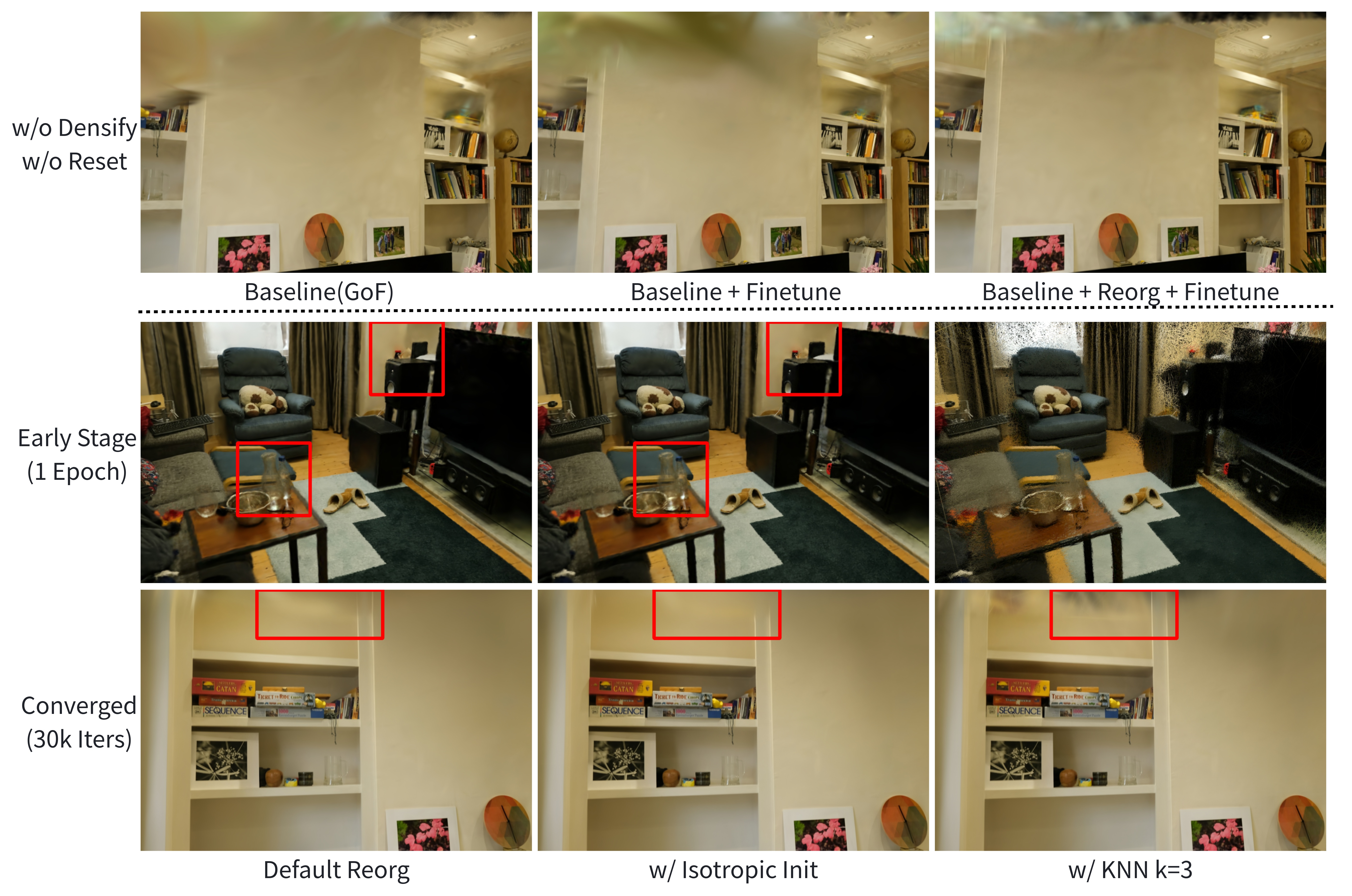}
  \caption{Ablation studies of ReorgGS. Top (Effectiveness of Reorg): We demonstrate the necessity of topological reorganization. We continue training the GoF baseline for 30k iterations without densification or opacity reset. Under this fixed topology, naive finetuning (middle) fails to recover sharp details (e.g., bookshelf) because the Gaussians are trapped in a deadlocked, suboptimal distribution. In contrast, applying our Reorg prior to the same finetuning (right) resolves the geometric occlusion, enabling pure gradient descent to successfully restore high-frequency structures.
Middle \& Bottom (Reorg Strategies): Ablation on Reorg specific designs across different stages (1 epoch vs. 30k iters). Discarding the covariance prior via Isotropic Initialization ("Iso-init") struggles to capture complex geometry early on (glass, middle row) and leaves splotchy noise on flat surfaces (wall, bottom row). Altering the splitting scale to $k=3$ leads to slower structural recovery. Our default Reorg strategy maintains optimal convergence and artifact-free rendering.}
  \label{fig:ablations}
\end{figure}

\section{Experiments}

\subsection{Experimental Setup}

Our experiments examine whether ReorgGS can convert a converged 3DGS model into an equivalent parameterization that is easier to optimize, without increasing the number of Gaussians. Each backbone is first trained to convergence with its original training procedure. We then compare three settings under the same additional optimization budget:

\begin{itemize}
\item \textbf{Baseline}: the converged original model, with no further modification.
\item \textbf{Direct continuation}: continued optimization of the original Gaussian set.
\item \textbf{ReorgGS}: equivalent distribution reorganization followed by continued optimization.
\end{itemize}

Unless otherwise stated, ReorgGS keeps exactly the same number of Gaussians as the model before reorganization. The experiments therefore compare parameterization structure, rather than model capacity or training budget.

\textbf{Backbones.}
We evaluate ReorgGS on vanilla 3DGS~\citep{kerbl20233d}, GOF~\citep{yu2024gaussian}, Taming-3DGS~\citep{mallick2024taming}, 3DGS-MCMC~\citep{kheradmand20243d}, and LiteGS~\citep{liao2025litegs}. These methods use different training mechanisms and Gaussian organizations, allowing us to test whether ReorgGS improves continued optimization across backbones.

\textbf{Datasets.} We use standard novel-view synthesis datasets, including Mip-NeRF 360~\citep{barron2022mip}, Blender~\citep{mildenhall2021nerf}, and Tanks\&Temples~\citep{Knapitsch2017tanks}, evaluating with PSNR, SSIM, and LPIPS~\citep{zhang2018perceptual}. 

\textbf{Gaussian-budget control.}
To control model capacity, we first train standard 3DGS to convergence and record the target point count. Other backbones reuse this scene-wise budget unless specified otherwise in point-budget ablations. We additionally consider LiteGS-half ($\times$0.5 budget) and LiteGS-double ($\times$2 budget) to test robustness. Further analyses on experimental setups, statistical variance, and computational costs are deferred to Appendix~\ref{app: exp_details}.
%

\subsection{Main Results: Equal-budget Comparison across Backbones}

Table~\ref{tab:main_results} compares the baseline, direct continuation, and ReorgGS under the same additional optimization budget. ReorgGS improves training PSNR for every backbone and improves test PSNR in most settings, with gains over direct continuation of ~0.5 dB on training views and ~0.14 dB on test views. This result shows that the plateau of a converged model is not caused only by insufficient iterations: under the same capacity and the same additional budget, reorganizing the parameterization reaches optimization regions that direct continuation cannot reliably access.

\begin{table}[thp]
\centering
\caption{Equal-budget comparison across backbones. Each cell reports train / test. ReorgGS keeps the same Gaussian count as the corresponding converged model.}
\label{tab:main_results}
\resizebox{\textwidth}{!}{%
\begin{tabular}{lccccccccc}
\toprule
& \multicolumn{3}{c}{PSNR} & \multicolumn{3}{c}{SSIM} & \multicolumn{3}{c}{LPIPS} \\
\cmidrule(lr){2-4} \cmidrule(lr){5-7} \cmidrule(lr){8-10}
Backbone & Baseline & Direct & ReorgGS & Baseline & Direct & ReorgGS & Baseline & Direct & ReorgGS \\
\midrule
3DGS & 31.37 / 28.26 & \cellcolor{pink!70}32.61 / 28.72 & \cellcolor{red!35}33.16 / 28.89 & 0.918 / 0.872 & \cellcolor{pink!70}0.936 / 0.880 & \cellcolor{red!35}0.940 / 0.883 & 0.169 / 0.198 & \cellcolor{pink!70}0.137 / 0.174 & \cellcolor{red!35}0.130 / 0.169 \\
GOF & 30.65 / 27.99 & \cellcolor{pink!70}32.76 / 28.80 & \cellcolor{red!35}33.21 / 28.93 & 0.916 / 0.873 & \cellcolor{pink!70}0.940 / 0.882 & \cellcolor{red!35}0.941 / 0.884 & 0.167 / 0.194 & \cellcolor{pink!70}0.131 / 0.170 & \cellcolor{red!35}0.130 / 0.169 \\
Taming-3DGS & 31.04 / 28.44 & \cellcolor{pink!70}32.72 / 28.84 & \cellcolor{red!35}33.38 / 29.08 & 0.911 / 0.874 & \cellcolor{pink!70}0.937 / 0.881 & \cellcolor{red!35}0.942 / 0.886 & 0.178 / 0.194 & \cellcolor{pink!70}0.135 / 0.171 & \cellcolor{red!35}0.129 / 0.166 \\
3DGS-MCMC & 31.69 / 28.71 & \cellcolor{pink!70}33.09 / 28.99 & \cellcolor{red!35}33.63 / 29.06 & 0.928 / 0.886 & \cellcolor{pink!70}0.944 / 0.883 & \cellcolor{red!35}0.946 / 0.885 & 0.151 / 0.177 & \cellcolor{pink!70}0.125 / 0.168 & \cellcolor{red!35}0.123 / 0.166 \\
LiteGS & 32.91 / 29.09 & \cellcolor{pink!70}33.15 / 28.95 & \cellcolor{red!35}33.67 / 29.03 & 0.938 / 0.888 & \cellcolor{pink!70}0.945 / 0.884 & \cellcolor{red!35}0.947 / 0.885 & 0.135 / 0.167 & \cellcolor{pink!70}0.123 / 0.165 & \cellcolor{red!35}0.121 / 0.165 \\
\hline
LiteGS-half & 32.12 / 28.87 & \cellcolor{pink!70}32.32 / 28.86 & \cellcolor{red!35}32.77 / 28.96 & 0.925 / 0.884 & \cellcolor{pink!70}0.932 / 0.882 & \cellcolor{red!35}0.934 / 0.883 & 0.156 / 0.180 & \cellcolor{red!35}0.144 / 0.174 & \cellcolor{pink!70}0.144 / 0.176 \\
LiteGS-double & 33.63 / 29.07 & \cellcolor{pink!70}33.91 / 28.95 & \cellcolor{red!35}34.47 / 28.97 & 0.948 / 0.888 & \cellcolor{pink!70}0.954 / 0.882 & \cellcolor{red!35}0.956 / 0.883 & 0.117 / 0.160 & \cellcolor{pink!70}0.105 / 0.160 & \cellcolor{red!35}0.103 / 0.159 \\
\bottomrule
\end{tabular}%
}
\end{table}


The training-view gains are consistently larger than the test-view gains, which is meaningful for scene-specific 3DGS because the training views are the  observations used to reconstruct each scene. Under a fixed Gaussian count and the same additional optimization budget, better training-view fitting indicates that ReorgGS uses capacity previously wasted by occlusion and coupling, while test-view metrics verify that the improvement does not collapse into view-specific artifacts. 


Fig.~\ref{fig:result} qualitatively supports the same trend: ReorgGS reduces topology-induced blur across backbones, and iterative reorganization further refines fine details. For more comprehensive qualitative and quantitative results across different datasets, please refer to Appendix~\ref{app: qc}.

The reorganized spatial structure also reduces redundant alpha compositing; in representative settings, ReorgGS provides up to 12.8\% rendering speedup.

\subsection{Robustness under Different Gaussian Budgets}

The LiteGS-half, LiteGS, and LiteGS-double rows in Table~\ref{tab:main_results} further vary the Gaussian budget while keeping the baseline, direct continuation, and ReorgGS at the same point count within each row. ReorgGS remains beneficial under both reduced and increased budgets. The improvement is larger under the half budget, where each Gaussian carries more representational pressure and a degenerate overlap pattern wastes a larger fraction of the available capacity. In this half-budget setting, Direct continuation slightly outperforms ReorgGS on LPIPS, likely due to the perceptual penalty of low-opacity initialization under severe capacity constraints, though ReorgGS still dominates in PSNR and SSIM. Under the double budget, the model has more redundancy and the test-view gain becomes smaller, but ReorgGS still improves training PSNR. This suggests that parameterization degeneration is not unique to compact models; it also appears when the Gaussian budget is larger, although its effect on novel-view metrics may be partly absorbed by the extra capacity.

\subsection{Cascaded Reorganization}

We further evaluate repeated Reorg operations. As shown in Table~\ref{tab:cascade}, Reorg x2, x3, and x4 continue to improve training PSNR, but the marginal gain decreases. This trend agrees with the degenerate-parameterization view: the first Reorg usually removes the strongest occlusion barriers and redundant coupling, while later reorganizations face fewer remaining structural obstacles.

To rule out the effect of total training budget, we compare two equal-budget trajectories. Direct continuation optimizes the original parameterization for the same total number of iterations, while Reorg $\times N$ allocates the same additional optimization budget after each reorganization. Table~\ref{tab:cascade} reports PSNR only and lists each backbone and point budget separately. Reorg remains better overall under the same total budget, showing that the cascaded gain comes from repeatedly replacing degenerate parameterizations rather than simply using more optimization steps. Training PSNR gains saturate as the number of reorganizations increases, while test-view improvements remain small but generally positive. This is consistent with our analysis: ReorgGS first improves optimization accessibility and local parameter decoupling, which is most directly reflected in training-view fitting. Test-view improvement then depends on whether the better fitting becomes a more consistent geometric and appearance representation.

\begin{table}[htp]
\centering
\caption{Cascaded reorganization under equal total optimization budget. Each cell reports train / test PSNR. For ReorgGS columns, darker color indicates a larger number of reorganization stages.}
\label{tab:cascade}
\resizebox{\textwidth}{!}{%
\begin{tabular}{lcccccc}
\toprule
& \multicolumn{2}{c}{2×iters} & \multicolumn{2}{c}{3×iters} & \multicolumn{2}{c}{4×iters} \\
\cmidrule(lr){2-3} \cmidrule(lr){4-5} \cmidrule(lr){6-7}
Backbone & Direct & ReorgGS & Direct & ReorgGS & Direct & ReorgGS \\
\midrule
3DGS & 33.07 / 28.77 & \cellcolor{red!15}33.46 / 28.90 & 33.26 / 28.80 & \cellcolor{red!25}33.57 / 28.93 & 33.42 / 28.79 & \cellcolor{red!35}33.62 / 28.93 \\
GOF & 33.10 / 28.89 & \cellcolor{red!15}33.54 / 28.93 & 33.29 / 28.92 & \cellcolor{red!25}33.59 / 28.97 & 33.41 / 28.93 & \cellcolor{red!35}33.66 / 28.97 \\
Taming-3DGS & 33.25 / 28.89 & \cellcolor{red!15}33.71 / 29.04 & 33.53 / 28.89 & \cellcolor{red!25}33.79 / 29.04 & 33.60 / 28.88 & \cellcolor{red!35}33.78 / 29.04 \\
3DGS-MCMC & 33.33 / 29.01 & \cellcolor{red!15}33.74 / 29.06 & 33.50 / 29.02 & \cellcolor{red!25}33.76 / 29.06 & 33.61 / 28.97 & \cellcolor{red!35}33.76 / 29.05 \\
LiteGS & 33.35 / 28.91 & \cellcolor{red!15}33.76 / 29.04 & 33.52 / 28.89 & \cellcolor{red!25}33.80 / 28.97 & 33.66 / 28.89 & \cellcolor{red!35}33.78 / 29.00 \\
LiteGS-half & 32.52 / 28.89 & \cellcolor{red!15}32.82 / 28.94 & 32.68 / 28.92 & \cellcolor{red!25}32.81 / 28.97 & 32.76 / 28.90 & \cellcolor{red!35}32.80 / 29.00 \\
LiteGS-double & 34.17 / 28.93 & \cellcolor{red!15}34.63 / 28.93 & 34.37 / 28.86 & \cellcolor{red!25}34.69 / 28.90 & 34.50 / 28.83 & \cellcolor{red!35}34.68 / 28.89 \\
\bottomrule
\end{tabular}%
}
\end{table}

\subsection{Necessity of Reorganization (w/o Densification \& Reset)}

To demonstrate that persistent floaters and blurry details are symptoms of topological deadlocks rather than mere lack of training, we evaluate finetuning under a strict constraint: disabling both heuristic densification and opacity reset. 

As shown in the top row of Fig.~\ref{fig:ablations}, directly finetuning the baseline under this fixed topology (Baseline + Finetune) fails to resolve the blurry artifacts on the bookshelf. The existing high-opacity floaters maintain their occlusion advantage, blocking gradients to the true surfaces and trapping the optimization in a local minimum. In contrast, applying ReorgGS before finetuning explicitly breaks these old visibility structures. By resampling the geometry with low opacity, ReorgGS smoothly eliminates gradient deadlocks, enabling pure gradient descent to successfully restore sharp, high-fidelity structures without relying on heuristic cleanup.

\subsection{Ablation on Reorg Configurations}

We further ablate the internal design choices of Reorg across different training stages, as visualized in the bottom two rows of Fig.~\ref{fig:ablations}. All variants use the same sampling distribution, differing only in the initialized state.

\textbf{Isotropic vs. Anisotropic Initialization.} 
We compare our default kNN-based anisotropic initialization against an isotropic variant (which serves as a minimalist implementation of Reorg without geometric priors). As observed in the early stage (1 epoch, left vs. middle), the isotropic variant struggles to capture structures like the glass bottle and the wall corner. While it eventually converges to a comparable visual quality at 30k iterations without statistically significant degradation, our default anisotropic initialization effectively aligns primitives to local manifolds much earlier, leading to fundamentally faster and more stable structural recovery.

\textbf{kNN Neighborhood Size ($k$).} 
We set $k=20$ as our default configuration to provide robust local geometric estimation. Shrinking the neighborhood size aggressively (e.g., "w/ kNN $k=3$") makes the covariance estimation highly susceptible to local noise. As shown in the right column of Fig.~\ref{fig:ablations}, this introduces severe needle-like noise artifacts during the early optimization stage. Although most of these spikes are eliminated upon convergence, noticeable artifacts still persist in poorly observed regions (e.g., the upper wall edge at 30k iterations). The default $k=20$ smoothly avoids these instabilities.

\section{Limitations and Broader Impact}

ReorgGS is a reorganization method rather than a geometry completion method. It assumes that the converged input model already contains reasonable scene support, and therefore cannot recover structures that are completely missing or never observed in the original Gaussian set. Its effectiveness may also decrease when the input distribution is dominated by severe reconstruction errors, since resampling from such a distribution can preserve part of the wrong support. In addition, ReorgGS requires additional optimization after reorganization.  ReorgGS can reduce persistent floaters and redundant overlap, which may improve the reliability and efficiency of scene-specific 3D reconstruction systems. At the same time, higher-quality reconstruction can increase privacy risks when applied to people, private spaces, or restricted environments. As with other high-fidelity 3D reconstruction methods, ReorgGS should be used only with appropriate consent, data-use permission, and care when releasing reconstructed scenes or trained Gaussian models.

\section{Conclusion}
We identify parameterization degeneration as a late-stage bottleneck in 3DGS: a converged model may represent the scene reasonably well while retaining floaters, redundant overlap, and poor gradient accessibility. ReorgGS addresses this by rebuilding the converged Gaussian set as an equivalent low-opacity, locally aligned parameterization, without changing the renderer or increasing capacity. Across backbones and Gaussian budgets, this reorganization improves fixed-budget optimization, suppresses persistent floaters, and reduces rendering redundancy, suggesting that the organization of Gaussian density, opacity, covariance, and visibility is a central factor in late-stage 3DGS performance.

\bibliographystyle{plainnat}
\bibliography{neurips}

\newpage
\appendix

\section{Theoretical proofs}
\label{app: tp}
\subsection{Distributional Equivalence and Constant-opacity Occupancy}
\label{app: deco}

Let the converged 3DGS model induce the normalized scene occupancy distribution

\[
p(x)=\frac{1}{Z}\sum_{i=1}^{N}\alpha_i\mathcal{N}(x\mid\mu_i,\Sigma_i),
\qquad
Z=\int\sum_{i=1}^{N}\alpha_i\mathcal{N}(x\mid\mu_i,\Sigma_i)dx.
\]

ReorgGS samples new centers from this distribution. Equivalently, it first selects an original Gaussian according to

\[
P(z_j=k)=\frac{\alpha_k}{\sum_{l=1}^{N}\alpha_l},
\]

and then draws a coordinate from

\[
x_j\sim\mathcal{N}(\mu_k,\Sigma_k).
\]

Given the resampled centers, ReorgGS estimates a local covariance \(\Sigma_j^q\) from the kNN neighborhood and constructs

\[
q_M(x)=\frac{1}{M}\sum_{j=1}^{M}\mathcal{N}(x\mid x_j,\Sigma_j^q).
\]

\paragraph{Proposition (distributional consistency).} Assume that \(p(x)\) is twice continuously differentiable in a local neighborhood, and that the kNN bandwidth satisfies \(k\to\infty\) and \(k/M\to0\). Then, as \(M\to\infty\),

\[
\mathbb{E}[q_M(x)]\to p(x),
\qquad
\mathrm{Var}[q_M(x)]=O(1/k).
\]

\paragraph{Proof.} For samples drawn from \(p\), the local 3D volume that contains \(k\) nearest neighbors satisfies

\[
V_k(x)\approx \frac{k}{Mp(x)}.
\]

The corresponding local covariance scale is therefore

\[
\|\Sigma^q(x)\|=O\left((k/M)^{2/3}\right).
\]

Treating the kNN local estimator as a kernel density estimator with bandwidth \(\Sigma^q(x)\), applying local Gaussian smoothing to \(p\), and using a second-order Taylor expansion gives

\[
\mathbb{E}[q_M(x)]
=
p(x)+\frac{1}{2}\mathrm{Tr}(\Sigma^q(x)\nabla^2p(x))+o(\|\Sigma^q(x)\|).
\]

Since \(k/M\to0\), the bandwidth vanishes and the bias term goes to zero. Hence \(\mathbb{E}[q_M(x)]\to p(x)\). For the variance of the local kernel estimator,

\[
\mathrm{Var}[q_M(x)]
\approx
\frac{1}{M}\int \mathcal{N}(x\mid z,\Sigma^q(z))^2p(z)dz.
\]

Using the squared Gaussian integral, the leading scale is

\[
\mathrm{Var}[q_M(x)]
\propto
\frac{p(x)}{M|\Sigma^q(x)|^{1/2}}.
\]

For a 3D kNN bandwidth, \(|\Sigma^q(x)|^{1/2}\propto k/M\). Thus,

\[
\mathrm{Var}[q_M(x)]\propto \frac{1}{k}.
\]

The variance therefore vanishes as \(k\to\infty\). This completes the proof.

This result shows that resampling and kNN covariance estimation preserve the original scene support in an asymptotic sense. ReorgGS further converts the original opacity weights into spatial sampling density and initializes the new primitives with a constant low opacity. Let the constant-opacity occupancy field after reorganization be

\[
\hat{F}(x)=\sum_{j=1}^{M}\alpha_0 K(x-x_j),
\]

where \(K\) is a local kernel. Since \(x_j\sim p(x)\),

\[
\mathbb{E}[\hat{F}(x)]
=
M\alpha_0\int K(x-z)p(z)dz.
\]

When the local kernel bandwidth is small,

\[
\mathbb{E}[\hat{F}(x)]\approx M\alpha_0 p(x).
\]

Thus, the occupancy strength originally represented by \(\alpha_i\) is represented after reorganization mainly by the sample density \(Mp(x)\), while each primitive starts from the same low opacity \(\alpha_0\). This separation is essential for the subsequent optimization analysis: ReorgGS preserves the distributional support while changing the parameterization among opacity, spatial density, and local covariance.

\subsection{Bounded Overlap under kNN Reorganization}
\label{app: bokr}

Distributional equivalence does not imply optimization equivalence. The original converged model may stack many high-opacity Gaussians in the same local region. These primitives make similar rendering contributions, but they also create strong overlap and strong parameter coupling. The kNN covariance in ReorgGS ties the support scale of each new Gaussian to the local sampling density, which controls the number of effective overlaps under standard local regularity assumptions.

Let the local density of reorganized centers be

\[
\rho(x)=Mp(x).
\]

Near \(x\), the volume that contains \(k\) nearest neighbors satisfies

\[
V_k(x)\approx \frac{k}{\rho(x)}.
\]

The kNN covariance makes the effective support volume of each new Gaussian proportional to \(V_k(x)\). Let \(K_q(x)\) denote the number of reorganized Gaussians that have a significant density or rendering response near \(x\). If the local density is smooth, the samples are sufficiently dense, and the kernel support is controlled by the kNN scale, then

\[
K_q(x)
\approx
\rho(x)\cdot V_k(x)
\approx
\rho(x)\cdot\frac{k}{\rho(x)}
=
O(k).
\]

This statement does not mean that the reorganized Gaussians do not overlap. Rather, it states that the number of effective local overlaps no longer grows without bound with the number of redundant Gaussians in a degenerate original cluster. Dense regions produce more centers but smaller covariances; sparse regions produce fewer centers but larger covariances. These two effects compensate each other, keeping the effective local coverage at a scale controlled by \(k\).

In contrast, standard densification clones or splits existing Gaussians near their current locations, and can preserve or amplify an existing local pile-up. If a degenerate cluster contains \(s\) strongly overlapping Gaussians, its local effective overlap can grow with \(s\). ReorgGS resamples centers from the global empirical distribution and re-estimates covariances, rewriting such local redundancy into a density-controlled representation with adaptive support. This geometric property underlies the reduction of alpha-compositing occlusion and Hessian / FIM coupling.

\subsection{Gradient Accessibility under Alpha Compositing}
\label{app: gaac}

For a single ray \(r\), the reconstruction loss is

\[
\mathcal{L}_r=\|C(r)-C^*(r)\|_2^2.
\]

For a responding primitive \(g\), ignoring higher-order occlusion interaction terms, the leading gradient term is scaled by its foreground transmittance:

\[
\|\nabla\mathcal{L}_r(g)\|
\propto
T_g(r)a(g,r)\|J_g(r)\|,
\]

where \(a(g,r)\) is the effective opacity of this primitive on the ray and \(J_g(r)\) is its local rendering Jacobian. When the effective foreground opacities are small,

\[
\log T_g(r)=\sum_{h<g}\log(1-a_h(r))\approx-\sum_{h<g}a_h(r).
\]

Defining

\[
A_g(r)=\sum_{h<g}a_h(r),
\]

we obtain

\[
T_g(r)\approx\exp(-A_g(r)).
\]

This gives the opacity barrier induced by alpha compositing: the larger the accumulated effective opacity in front of a target surface, the smaller the gradient received by parameters near that surface. For high-opacity floaters in a degenerate distribution \(p\), the true behavior is more severe than the small-opacity approximation. If a foreground Gaussian satisfies

\[
a_j^p(r)\to1,
\]

then

\[
\log(1-a_j^p(r))\to-\infty,
\qquad
T_g^p(r)=\prod_{h<g}(1-a_h^p(r))\to0.
\]

Thus, a high-opacity floater can approximately cut off the optimization path to the true surface behind it.

Since ReorgGS changes the Gaussian set, there is no strict index-wise correspondence between the primitives before and after reorganization. We therefore consider the same target physical surface near depth \(d\) along ray \(r\), and compare the responding primitives \(g_p\) and \(g_q\) under the original and reorganized parameterizations. Let \(A_d^p(r)\) be the accumulated effective opacity in front of the target surface in the original distribution. In the reorganized distribution, if there are at most \(m_q(r,d)\) effective responding primitives in front of the surface and each satisfies \(a_j^q(r)\le\alpha_0\), then

\[
A_d^q(r)=\sum_{j<d}a_j^q(r)\le m_q(r,d)\alpha_0,
\]

and

\[
T_d^q(r)=\prod_{j<d}(1-a_j^q(r))
\ge
(1-\alpha_0)^{m_q(r,d)}
\approx
\exp(-m_q(r,d)\alpha_0).
\]

\paragraph{Proposition (gradient accessibility).} If \(A_d^q(r)\le m_q\alpha_0\) and the original degenerate distribution satisfies \(A_d^p(r)>m_q\alpha_0\), then

\[
T_d^q(r)>T_d^p(r).
\]

Furthermore, for the responding primitives \(g_q\) and \(g_p\) near the same target physical surface,

\[
\frac{\|\nabla \mathcal{L}_r(g_q)\|}{\|\nabla \mathcal{L}_r(g_p)\|}
\approx
\exp(A_d^p(r)-A_d^q(r))
\cdot
\frac{a(g_q,r)}{a(g_p,r)}.
\]

\paragraph{Proof.} By alpha compositing,

\[
T_d^p(r)\approx\exp(-A_d^p(r)),
\qquad
T_d^q(r)\approx\exp(-A_d^q(r)).
\]

Together with \(A_d^q(r)\le m_q\alpha_0<A_d^p(r)\), this gives

\[
T_d^q(r)
\ge
\exp(-m_q\alpha_0)
>
\exp(-A_d^p(r))
\approx
T_d^p(r).
\]

For any responding primitive \(g\), the leading gradient term satisfies

\[
\|\nabla\mathcal{L}_r(g)\|
\propto
T_d(r)a(g,r)\|J_g(r)\|.
\]

Since \(g_p\) and \(g_q\) respond to the same local physical surface, their Jacobian norms are of the same order. Hence,

\[
\frac{\|\nabla \mathcal{L}_r(g_q)\|}{\|\nabla \mathcal{L}_r(g_p)\|}
\approx
\frac{T_d^q(r)a(g_q,r)}{T_d^p(r)a(g_p,r)}
\approx
\exp(A_d^p(r)-A_d^q(r))
\cdot
\frac{a(g_q,r)}{a(g_p,r)}.
\]

This completes the proof.

The expression also explains the trade-off introduced by low-opacity initialization. Because \(\alpha_0\) is small, the local factor \(a(g_q,r)/a(g_p,r)\) may be below one. However, \(A_d^p(r)\) is a macroscopic occlusion quantity accumulated over all foreground floaters and redundant Gaussians along the ray, and it affects transmittance exponentially. When the original parameterization contains a strong opacity barrier, removing that barrier can yield an exponential recovery of gradients that dominates the local linear reduction caused by low opacity.

\subsection{Hessian / FIM Coupling and Local Manifold Alignment}
\label{app: hflma}

Gradient accessibility explains how ReorgGS reopens optimization paths to occluded surfaces. A second degeneracy comes from parameter coupling. If multiple Gaussians strongly overlap and explain the same pixel residuals, their Jacobian columns become highly correlated, making it difficult for the optimizer to determine which parameter block should be updated.

Define the opacity-weighted overlap energy as

\[
\mathcal{O}(\mathcal{G})
=
\sum_{i\neq j}\alpha_i\alpha_j\int G_i(x)G_j(x)dx.
\]

Assuming normalized Gaussians,

\[
\int \mathcal{N}(x\mid\mu_i,\Sigma_i)\mathcal{N}(x\mid\mu_j,\Sigma_j)dx
=
\mathcal{N}(\mu_i\mid\mu_j,\Sigma_i+\Sigma_j).
\]

Thus, \(\mathcal{O}(\mathcal{G})\) measures the opacity-weighted strength of spatial overlap. It can also be viewed as a geometric proxy for local optimization coupling. Let all parameters be \(\Theta=\{\theta_i\}\). Under the Gauss--Newton approximation,

\[
H\approx J^TJ,
\]

where \(J=\partial C/\partial\Theta\). The off-diagonal block for parameter blocks \(i,j\) is

\[
H_{ij}
\approx
\left(\frac{\partial C}{\partial\theta_i}\right)^T
\left(\frac{\partial C}{\partial\theta_j}\right).
\]

The off-diagonal blocks of the Fisher Information Matrix are similarly determined by inner products between the effects of different parameters on the observation distribution. If two Gaussians strongly overlap in space, they affect many of the same pixels and rays, making their Jacobian columns highly correlated. Therefore, \(\int G_iG_j\) is not only a measure of geometric redundancy, but also a proxy for off-diagonal coupling in the Hessian / FIM.

\paragraph{Proposition (overlap energy reduction).} In a degenerate distribution \(p\), if a local cluster contains \(s\) strongly overlapping Gaussians with average opacity \(\bar{\alpha}\), the overlap energy grows approximately quadratically:

\[
\mathcal{O}_p=\Omega(s^2\bar{\alpha}^2).
\]

This quadratic growth corresponds to severe parameter multicollinearity. If only a standard opacity reset is applied, then

\[
\mathcal{O}_{\mathrm{reset}}
=
\alpha_0^2\sum_{i\neq j}\int G_i^p(x)G_j^p(x)dx.
\]

It reduces the opacity weights, but preserves the geometric overlap term of the old parameterization. After ReorgGS, we instead have

\[
\mathcal{O}_{\mathrm{Reorg}}
=
\alpha_0^2\sum_{i\neq j}\int G_i^q(x)G_j^q(x)dx.
\]

By the bounded-overlap analysis in Section~\ref{app: bokr}, if the effective local overlap after reorganization is controlled by \(m_q\), then

\[
\mathcal{O}_q=O(Mm_q\alpha_0^2),
\]

whereas the original overlap in a degenerate cluster can reach \(\Omega(s^2\bar{\alpha}^2)\). ReorgGS therefore reduces not only the opacity weights, but also the geometric overlap induced by the parameterization. This weakens off-diagonal Hessian / FIM coupling and makes the local curvature closer to a block-diagonal structure.

Finally, the kNN covariance provides a local manifold-alignment effect. If the scene is locally approximated by a low-dimensional manifold, let \(U\) denote the principal directions of the local neighborhood and \(\Lambda\) its variance spectrum. The kNN covariance estimate gives

\[
\Sigma_j^q\approx U\Lambda U^T.
\]

Since the local curvature of Gaussian position and scale parameters is related to \((\Sigma_j^q)^{-1}\), the aggregate local curvature of several reorganized Gaussians can be approximated as

\[
H_{\mathrm{local}}
\propto
\sum_j(\Sigma_j^q)^{-1}
\approx
K\,U\Lambda^{-1}U^T.
\]

where \(K\) denotes the number of reorganized Gaussians in the local neighborhood. This does not mean that ReorgGS exactly aligns with the global eigenvectors of the Hessian, nor that it constructs an optimal preconditioner. It means that kNN covariance estimation makes the principal directions of local Gaussians consistent with the sampled geometric neighborhood, reducing conflicting directional updates. Together with low-opacity initialization and bounded overlap, ReorgGS converts a degenerate parameterization caused by high opacity, redundant stacking, and inconsistent orientations into a distributionally similar but more gradient-accessible and less coupled one.

\section{Additional Experimental Details and Discussions}
\label{app: exp_details}

In this section, we provide further context on our experimental design, addressing potential questions regarding the opacity reset baseline, statistical variance in our sampling process, and a detailed breakdown of computational costs. Our implementation will be released as open-source code. All experiments are run on NVIDIA A100 GPUs.

\subsection{Computational Cost: Training Overhead and Rendering Speedup}
While ReorgGS explicitly improves the rendering speed, we also transparently detail its computational overhead during the post-training phase.

\textbf{Training \& Module Overhead.} As a post-training method, ReorgGS inherently increases the total training time due to the additional gradient descent steps required. For instance, each post-training stage adds approximately $5$ minutes for the LiteGS architecture on an NVIDIA A100 GPU. However, the ReorgGS algorithmic module itself is extremely lightweight. Leveraging efficient GPU kNN implementations, the complete topological reorganization (resampling and covariance re-estimation) of millions of Gaussians takes only a few seconds. Since this module is invoked very infrequently (e.g., 1 to 3 times per scene), its core algorithmic overhead is completely negligible compared to the rendering backward pass.

\textbf{Rendering Speedup.} Crucially, although the Gaussian budget is strictly matched to the baseline count ($M=N$), the reorganized topology inherently reduces rendering redundancy. By explicitly bounding local geometric overlap via K-NN initialization, fewer effective Gaussians participate in the depth-sorting and alpha-compositing operations per pixel tile. Validated on the LiteGS architecture, ReorgGS successfully reduces the total inference time from $10.75$ seconds down to $9.37$ seconds, yielding a concrete $12.8\%$ acceleration without compromising geometric fidelity.

\subsection{On the Insufficiency of Pure Opacity Reset}
A natural question is why we do not isolate ``pure opacity reset'' as a separate comparative baseline to prove the effectiveness of ReorgGS. Importantly, in our experimental setup, the \textbf{Direct Continuation baseline already incorporates standard periodic opacity resets} (as per the default 3DGS training schedule). The fact that direct continuation still plateaus with blurry artifacts proves that merely resetting opacity on a degenerate topology is fundamentally insufficient.

The operations are simply not on the same level. Opacity reset only scales down the edge weights of the existing overlap graph ($\alpha_i \leftarrow \alpha_0$). The overlapping primitives maintain the exact same centers $\mu$ and covariance matrices $\Sigma$, meaning their Jacobians remain highly collinear. To rapidly minimize the photometric loss, the optimizer quickly scales the opacities of these exact same redundant Gaussians back up, returning the model to its original topological deadlocks. In contrast, ReorgGS operates on a fundamentally deeper structural level. By probabilistically resampling centers and re-estimating kNN covariances, it physically rebuilds the overlap graph and fundamentally breaks the parameter coupling.

\subsection{Robustness to Random Seeds (Statistical Stability)}
Because ReorgGS involves categorical sampling and Gaussian coordinate sampling, one might ask whether the choice of random seed introduces high variance into the final reconstruction, and whether averaging results across multiple seeds per scene is necessary.

In practice, this statistical variance is entirely marginalized out by two macro-level factors. First, at the \emph{intra-scene} level, a typical 3DGS model contains millions of primitives ($M \approx 10^5 \sim 10^6$). Drawing such a massive number of independent samples ensures that the macroscopic occupancy distribution tightly converges to the empirical field $p(x)$ (as formalized in Proposition A.1, $\mathrm{Var}[q_M(x)] \propto 1/k \to 0$). Second, at the \emph{inter-scene} level, our quantitative tables report metrics averaged across dozens of diverse scenes from multiple datasets. This combination of massive large-scale sampling within each scene and macro-averaging across multiple datasets completely flattens any microscopic statistical fluctuations caused by individual random seeds, making the reported performance highly stable and reproducible.

\section{Qualitative Comparisons}
\label{app: qc}
\begin{figure}[h!]
  \centering 
    \includegraphics[width=1.0\linewidth]{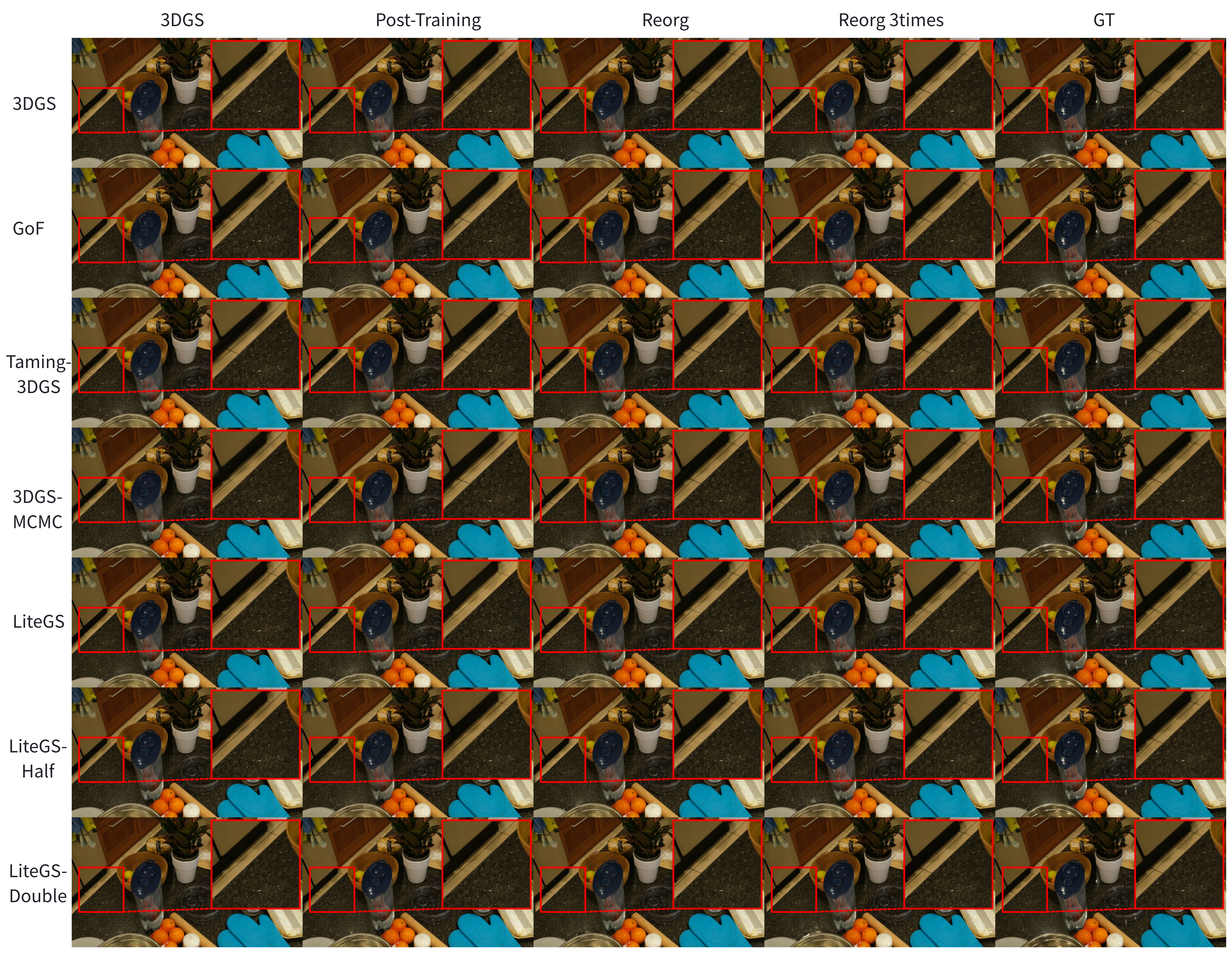}
  \caption{\textbf{Qualitative comparisons on the Kitchen scene.} We evaluate Reorg across six baselines (rows). Columns show the original baseline, standard post-training, Reorg (1 pass), Reorg (3 passes), and Ground Truth. As highlighted in the red insets, standard post-training struggles with the dark bottle boundaries and countertop reflections, whereas Reorg effectively breaks the optimization deadlocks to restore sharp, high-fidelity details.}
  \label{fig:kitchen}
\end{figure}

As visualized in our qualitative results, ReorgGS consistently recovers finer geometric and textural details that are otherwise lost in baseline methods due to optimization deadlocks. We highlight specific challenging regions across different scenarios:

\textbf{KITCHEN (High-Frequency Textures):} For indoor scenes rich in fine-grained patterns, as shown in Fig.~\ref{fig:kitchen}, such as the floor tiles in the KITCHEN dataset, direct finetuning often yields over-smoothed results because redundant Gaussians blur the structural boundaries. ReorgGS redistributes the primitives to align with local manifolds, faithfully restoring the sharp, intricate textural details of the floor.

\begin{figure}[h!]
  \centering 
    \includegraphics[width=1.0\linewidth]{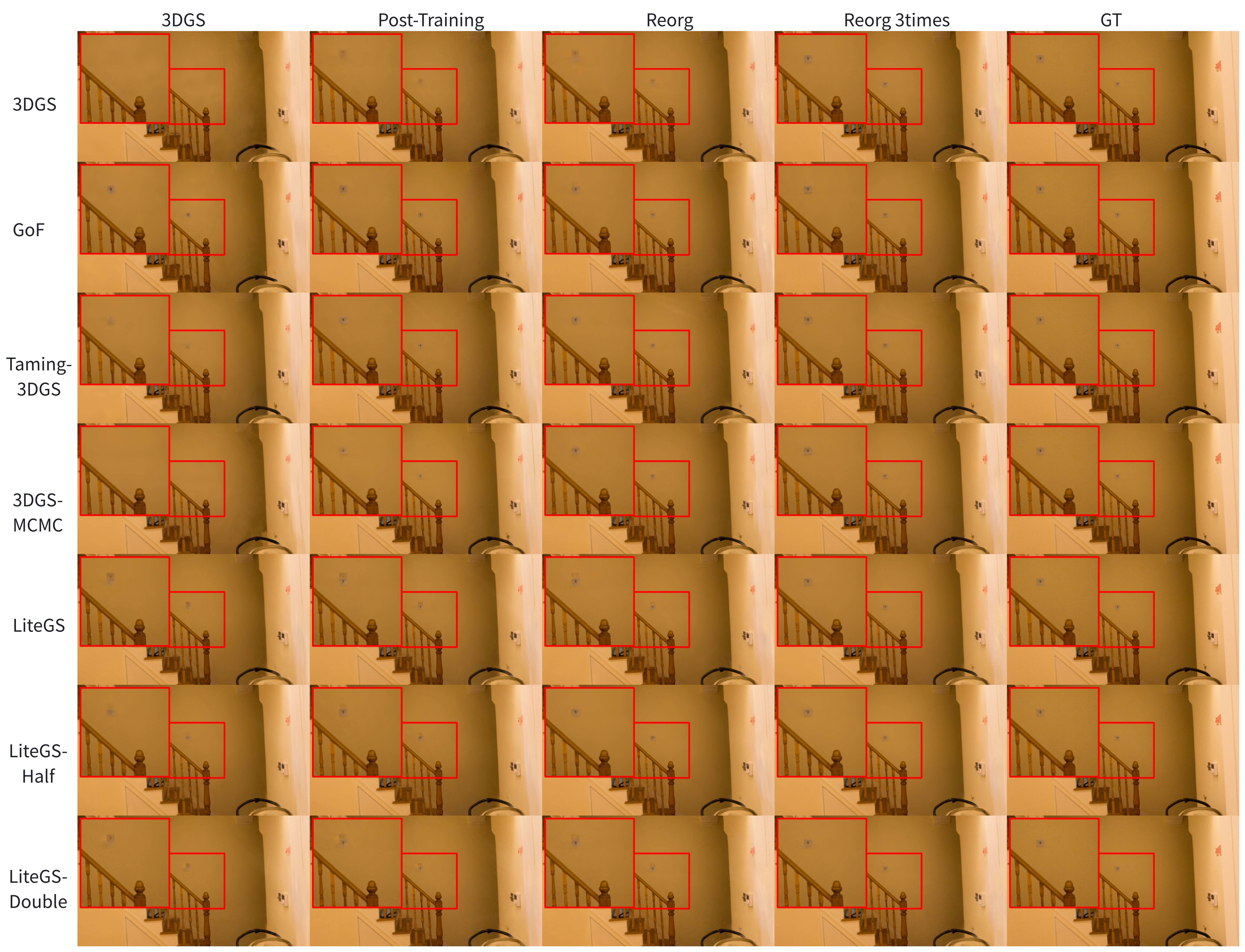}
  \caption{\textbf{Qualitative comparisons on the Playroom scene.} This scene challenges the parameterization of thin geometric structures (wooden railings) against a plain background. While standard post-training often leaves floating artifacts or broken geometry, Reorg successfully restores continuous and clean foreground structures without polluting the background.}
  \label{fig:playroom}
\end{figure}

\textbf{PLAYROOM (Ghosting Artifacts):} As shown in Fig.~\ref{fig:playroom}, baseline methods typically struggle with severe topological entanglement, which manifests as noticeable ghosting artifacts (double images) around the wall switches. By explicitly breaking these fixed occlusion structures, ReorgGS successfully converges to a single, crisp geometry for the switch. 

\begin{figure}[h!]
  \centering 
    \includegraphics[width=1.0\linewidth]{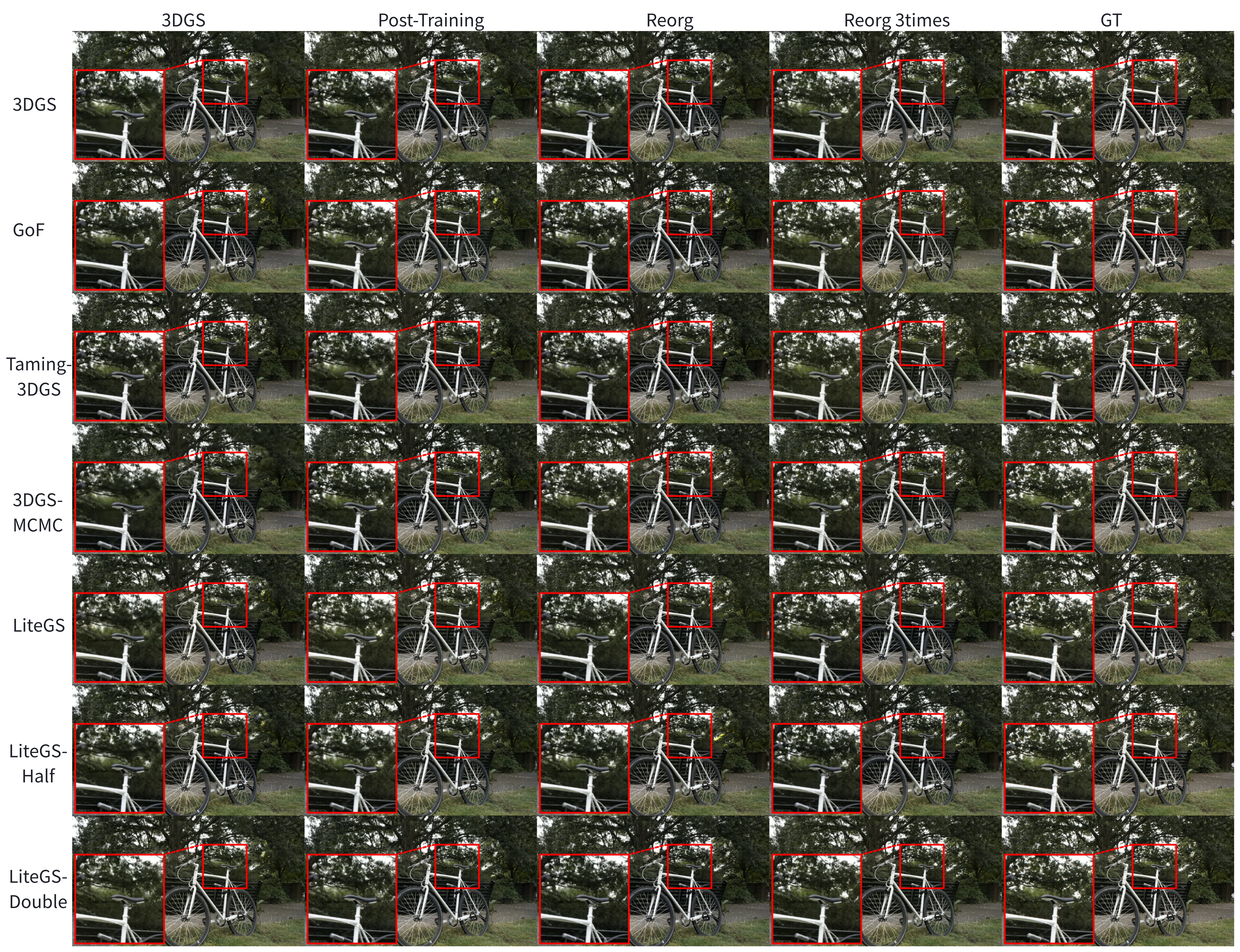}
  \caption{\textbf{Qualitative comparisons on the Bicycle scene.} This scene presents extreme depth occlusions between thin metallic cables and a high-frequency foliage background. Reorg accurately disentangles the foreground objects from the complex background, significantly alleviating the blurring artifacts seen in standard gradient-based finetuning.}
  \label{fig:bicycle}
\end{figure}

\textbf{BICYCLE (Complex Background Geometry):} In outdoor environments like the BIKE scene, as shown in Fig.~\ref{fig:bicycle}, recovering complex, depth-varying background structures is highly challenging. While baselines tend to merge the background leaves and branches into cloudy, semi-transparent artifacts, ReorgGS excels at resolving the crisp silhouettes and natural gaps of the intricate foliage, demonstrating its superior capability in modeling high-frequency geometric boundaries.

\newpage

\newpage
\newpage

\end{document}